\documentclass[11pt,letter,twoside]{rho-class/rho}
\usepackage{graphicx} 
\usepackage{xcolor,soul} 
\usepackage{subcaption}
\usepackage{authblk} 

\DeclareCaptionStyle{VariableCaptionStyle}[justification=centering]{justification=justified, margin=1em,labelfont={bf,sf},list=no,font=footnotesize}
\newcommand\clearfix{\vspace{-15pt}} 

\newcommand\blfootnote[1]{%
  \begingroup
  \renewcommand\thefootnote{}\footnote{#1}%
  \addtocounter{footnote}{-1}%
  \endgroup
}

\title{Learning, Potential, and Retention: An Approach for Evaluating Adaptive AI-Enabled Medical Devices}

\author[]{Alexis Burgon}
\author[]{Berkman Sahiner}
\author[]{Nicholas A Petrick}
\author[]{Gene Pennello}
\author[]{Ravi K Samala}

\affil[]{Division of Imaging, Diagnostics, and Software Reliability (DIDSR)

Office of Science and Engineering Laboratories (OSEL), 

Center for Devices and Radiological Health (CDRH)  

U.S. Food and Drug Administration 

10903 New Hampshire Ave, Silver Spring, MD US 20993}
\date{} 

\corres{Ravi K Samala}
\email{Ravi.Samala@fda.hhs.gov.}

\setbool{rho-abstract}{true} 
\setbool{corres-info}{true} 

\begin{abstract}
This work addresses challenges in evaluating \textit{adaptive} artificial intelligence (AI) models for medical devices, where iterative updates to both models and evaluation datasets complicate performance assessment. We introduce a novel approach with three complementary measurements: \textit{learning} (model improvement on current data), \textit{potential} (dataset-driven performance shifts), and \textit{retention} (knowledge preservation across modification steps), to disentangle performance changes caused by model adaptations versus dynamic environments. Case studies using simulated population shifts demonstrate the approach’s utility: gradual transitions enable stable learning and retention, while rapid shifts reveal trade-offs between plasticity and stability. These measurements provide practical insights for regulatory science, enabling rigorous assessment of the safety and effectiveness of adaptive AI systems over sequential modifications.    
\end{abstract}

\keywords{regulatory science, adaptive AI, artificial intelligence-enabled software as a medical device}

\begin{document}
\maketitle
\thispagestyle{firststyle}

\section{Introduction}
\blfootnote{Code available at \href{https://github.com/DIDSR/VIGILANT}{github.com/DIDSR/VIGILANT}.}
Artificial intelligence (AI)-enabled devices deployed in clinical settings traditionally use a ``locked" model, defined  as ``an algorithm that provides the same result each time the same input is applied to it and does not change with use" \cite{FDApccpRF}, \cite{FDAglossary}. The use of locked models is typically intended to prevent the emergence of unevaluated behaviors due to changes in system parameters. However, utilization of a locked model may not guarantee ongoing predictable performance. When applied within dynamic environments, such as those in clinical settings, locked models may experience performance decay over time \cite{davis2017calibration, chen2017decaying, nestor2019feature, sahiner2023data}. In contrast, continual learning \cite{de2021continual} AI models may be able to adapt to changing environments, though at the cost of potentially developing unexpected behaviors \cite{babic2019algorithms}.

\begin{figure}
\centering
\includegraphics[width=0.9\linewidth]{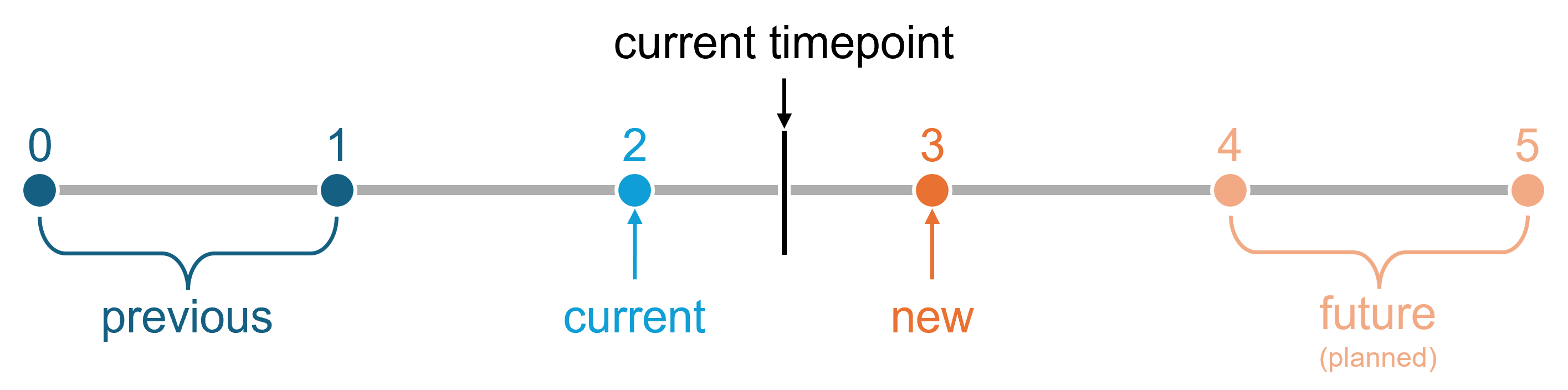}
\caption{A simple adaptation timeline for an adaptive AI system with an initial implementation (timepoint 0) and five modification steps (timepoints 1-5), showing the transition between timepoints 2 (current) and 3 (new).}
\label{fig:timeline}
\clearfix
\end{figure}

Adaptive AI models, a type of continual learning AI model that is versioned over time \cite{FDAglossary, IMDRFmlmdDefs}, present a potential middle ground between entirely static ``locked" models and those that continuously learn. In contrast to continually learning AI models, which view provided data as a constant stream for training and update with each sample or task encountered, adaptive AI is trained in distinct stages (modification steps), as shown in Figure \ref{fig:timeline}. These modification steps are typically triggered by a known change in the input, an observed decrease in performance, the accumulation of a large amount of new training data, or the transition to new technological advances. Furthermore, while modification steps indicate model changes, the overall \textit{task} (e.g., disease classification from chest X-rays) remains the same across all modification steps, in contrast to some continual learning models that learn sequences of different tasks \cite{lopez2017gradient}. This training approach enables model adaptation over time while also providing clear change points for evaluation. 

However, the evaluation of adaptive models is more complicated than that of their locked counterparts, largely due to the potentially variable nature of both the model and the evaluation data, which provide two axes for assessing change. If new modification steps of the model and dataset are created synchronously, the challenge is to determine the source of the change observed from one modification step to the next. \textit{Is an observed improvement in performance actually due to a change in the model, or is the new test dataset simply different?} 

A transition to an easier test dataset can result in overly optimistic performance evaluation, which may promote unwarranted trust in the model. Furthermore, if this change in the test dataset is accompanied by a similar change in the model's training dataset, there is the potential for overfitting to the new training data. This may result in a decrease in model performance when applied in a clinical setting, despite the measured increase in performance during evaluation.

Evaluation of adaptive AI is further complicated by the inherent trade-off between a model's \textit{plasticity} (ability to learn new information) and \textit{stability} (capacity to retain prior knowledge) \cite{mermillod2013stability}. The relative importance of plasticity and stability depends on the context in which an AI model is developed and deployed. For models that are modified infrequently using large, diverse datasets, plasticity may be prioritized over stability, as each new dataset is sufficiently representative of the intended population at that point in time. In contrast, prioritizing plasticity over stability in frequently updated models could be detrimental, since over-adaptation to changing populations may result in unpredictable model behavior. 

To provide insight into the complexities of adaptive AI evaluation, we propose and assess a novel approach consisting of three complementary measurements designed to provide a granular and insightful analysis of AI models across sequential modification steps. This approach provides diagnostic insight by measuring: (1) the model's \textit{learning} -- irrespective of changes to the evaluation data, (2) the \textit{potential} improvement available through adaptation to dataset changes, and (3) the model's \textit{retention} of knowledge learned in previous modification steps. Together, these measurements may provide insights into model plasticity and stability that extend beyond traditional performance assessment.

A preliminary version of this work was presented at the Radiological Society of North America (RSNA) 2023 Annual Meeting \cite{burgon2023}.

\section{Materials and Methods}

\subsection{Measurements}
Consider a model $M$ that is sequentially updated with different modification steps $V$. The corresponding evaluation dataset $D$ is likewise updated over time to ensure adaptation to changing populations and avoid over-fitting to a fixed dataset. These dataset updates may entail gathering a completely new dataset, or updating the previous step's data with newly acquired samples and/or the removal of old samples. The performance of model $M$ measured on dataset $D$ -- measured using any task-appropriate performance metric -- can be denoted $S(M|D)$. 

Although the concurrently updating model and dataset allow adaptation to changing populations, such evaluation overlooks nuances that may provide valuable insight into model changes over time, specifically the separation of the effects of the changing model and changing dataset. Independent evaluation at each modification step $V$ provides no indication of whether changes in performance are due to recent model adaptations or to differences in dataset difficulty.  This creates a challenge in the adaptive AI paradigm, wherein each model's performance is compared to the prior modification step. Updating datasets between steps provides a somewhat weaker basis for comparison, whereas using a fixed dataset across multiple steps increases the likelihood of model overfitting. To address this limitation, we present three measurements: (1) \textit{learning}, (2) \textit{potential} and (3) \textit{retention}. Each measurement provides different insight into the performance of a sequentially modified model. 

\begin{figure}[tb] 
    \centering
    \includegraphics[width=0.9\linewidth]{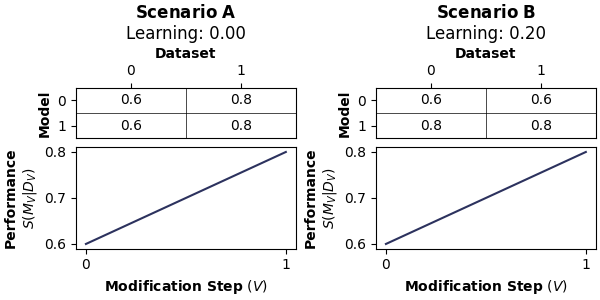}
    \caption{Comparison of \textit{learning} between Scenario A \& Scenario B using a toy example. Despite showing the same performance at each comparable modification step (0 \& 1), as indicated by their identical line plots, the two scenarios exhibit different \textit{learning}. The performance change in Scenario A is due to a shift in dataset difficulty, whereas the performance change in  Scenario B is due to an improvement in model knowledge. }
    \label{fig:ex-learning}
    \clearfix
\end{figure}

The first measurement, \textit{learning}  (equation \ref{eq:learning}) captures the difference in performance  due to changes in the model, and is measured with respect to the current dataset.
\begin{equation}\label{eq:learning}
    learning(M_V|D_V) = S(M_V|D_V) - S(M_{V-1}|D_V)
\end{equation}
 A toy example demonstrating the importance of the \textit{learning} measurement is shown in Figure \ref{fig:ex-learning} for different models (Scenarios A \& B). Both scenarios feature an initial model \& dataset (modification step 0) and the resulting model \& dataset following modification (modification step 1). When a straightforward performance assessment method is applied -- measuring each model's performance on its respective dataset -- the performance improvement appears identical in both scenarios: an increase from a performance of 0.6 to 0.8. However, an important discrepancy can be observed through inspection of the \textit{learning} measurement, as well as examining the pairwise performance between model and dataset modification steps (shown at the top of Figure \ref{fig:ex-learning}). The performance increase in Scenario A is entirely due to a change in dataset difficulty (learning = 0), while the increase in Scenario B  is entirely due to a change in model knowledge.

\begin{figure}[tb]
    \centering
    \includegraphics[width=0.9\linewidth]{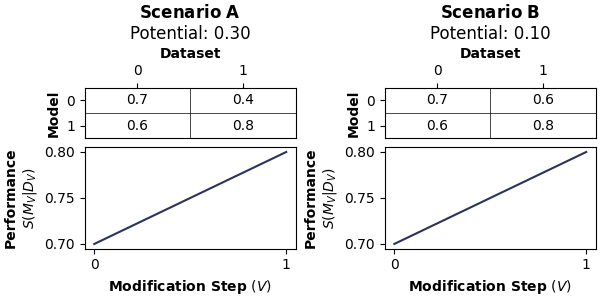}
    \caption{Comparison of \textit{potential} between Scenario A \& Scenario B using a toy example. Despite showing the same performance at each comparable modification step (0 \& 1), as indicated by their identical line plots, the two scenarios exhibit different \textit{potential}.  Scenario A demonstrates greater \textit{potential} because the modification step 1 dataset presented a greater challenge to the modification step 0 model than was observed in Scenario B. }
    \label{fig:ex-potential}
    \clearfix
\end{figure}

\textit{Potential}, as defined in equation \ref{eq:potential}, measures the change in performance resulting solely from differences in dataset challenge (measured with respect to the prior version of the model); i.e., the potential performance change that would have been observed had the model not been updated at all. \textit{Potential} provides a point of reference for the value of a modification to the model at step $V$.
Figure \ref{fig:ex-potential} presents a toy example of scenarios with similar performances $S(M_v|D_v), v=0,1$, but different \textit{potential} values. Although both examples show a performance increase of 0.1, only in the example on the right does the performance increase align with the \textit{potential} learning from the dataset difficulty changes.

\begin{equation} \label{eq:potential}
    potential(M_V|D_V) = S(M_{V-1}|D_{V-1}) - S(M_{V-1}|D_V) 
\end{equation}

\begin{figure}[htb]
    \centering
    \includegraphics[width=0.9\linewidth]{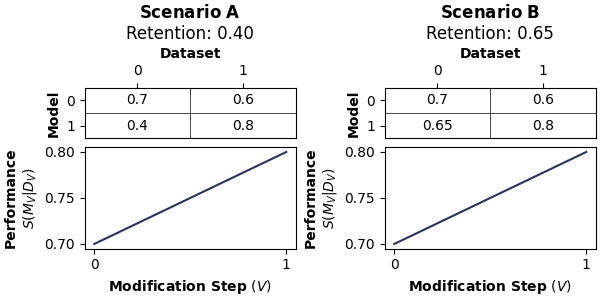}
    \caption{Comparison of \textit{retention} between Scenario A \& Scenario B using a toy example. Despite showing the same performance at comparable modification step (0 \& 1), as indicated by their identical line plots, the two scenarios exhibit different \textit{retention}.  Scenario A shows lower \textit{retention} because the modification step 1 model demonstrates greater performance degradation on the modification step 0 evaluation dataset.}
    \label{fig:ex-retention}
    \clearfix
\end{figure}

The final measurement, \textit{retention} (equation \ref{eq:retention}), assesses the model's preservation of knowledge evaluated at previous modification steps $(v \in [0...V-1])$. 

\begin{equation} \label{eq:retention}
    retention(M_V | D_V) = \sum_{v=0}^{V-1}S(M_V|D_v)\times W((V-1)-v)
\end{equation}
A weight term $W$ is introduced to account for the gradual deprecation of data over time (e.g., due to changing populations or clinical practices). $W$ is an exponential decay term $W(t)=e^{-\lambda t}$ , in our current approach, normalized such that $\sum_{v=1}^{V-1}{W((V-1)-v)} = 1$. The weight decay constant $\lambda$ may be tuned to adjust the rate of data deprecation as appropriate for a given use case. Throughout this work, we use a value of $\lambda= 0.5$.  Likewise, the decay term $W$ might be changed to account for other potential changes in the input data over time, such as a sudden shift in disease prevalence or the emergence of a new condition.   
A toy example demonstrating the utility of the \textit{retention} measurement is shown in Figure \ref{fig:ex-retention}. While both scenarios achieve similar performances, Scenario A on the left exhibits low \textit{retention} (as can be observed in the performance drop of model 1 when assessed on dataset 0). The \textit{retention} measurement adjusts the measured performance to account for consistency in model performance on previous datasets. The importance of knowledge retention depends on the models context of use and whether the populations represented in the previous modification steps still represent portions of the device's intended population \cite{chen2017decaying}.

\subsection{Case study}\label{sec:casestudy}


We demonstrate the utility of our proposed measurements in two simulated population shift scenarios: \textit{single} and \textit{double} population shift. To further highlight the nuances of these measurements, an additional variant of the single population shift scenario, referred to as "limited plasticity", was performed in which the learning capacity of the models used was limited.  In this variant, all but the final layer in the model were frozen for modifications after the initial step 0 model. This resulted in three experiments: (i) single population shift, (ii) single population shift -- limited plasticity, and (iii) double population shift. All experiments encompass an initial model \& dataset (step 0) and four subsequent modifications to each (step 1 to 4).
All models utilized the ResNet-18 architecture, with ImageNet-pretrained weights for a classification task. The data were from MIDRC's Open-A1 and Open-R1 repositories. ``Populations" are defined by the combination of source repository (Open-A1 vs Open-R1) and image acquisition method (CR versus DX X-rays). 
The population distributions at each modification step are shown in Figures \ref{fig:2p-data} and \ref{fig:3p-data} for the single and double population shift scenarios, respectively. Population distributions are kept constant across training, validation, and testing datasets. Both population shifts are gradual, with each modification step's dataset containing a subset of the previous step's dataset.
All datasets were stratified to include equal numbers of disease-positive and disease-negative patients to avoid performance changes resulting from changing prevalence, as summarized in Table \ref{tab:data_size}. The number of patients in each partition was consistent across modification steps and experiments.
Training and tuning of the AI model for the imaging classification task are described in detail elsewhere \cite{burgon2024decision,burgon2024bias}. For all three experiments, model performance was measured using the area under the receiver operating characteristic curve (AUROC). All results represent the mean and 95\% confidence interval from models across 25 repetitions.

\begin{table}[htb]
    \centering
    
    \begin{tabular}{rccc} 
         & \textbf{Training} & \textbf{Validation} & \textbf{Testing} \\ \hline
         \textbf{Disease Positive} & 525 & 75 & 150 \\ \hline 
         \textbf{Disease Negative} & 525 & 75 & 150 \\  \hline
         \textbf{Total} & \textbf{1050} & \textbf{150} & \textbf{300}
    \end{tabular}
    \caption{Number of patients included in each data partition (training, validation, testing) for each modification step. Images from a single patient were not split across different partitions.}
    \label{tab:data_size}
\end{table}

\section{Results}\label{sec:results}

\begin{figure}[tb]
    \centering
    \begin{subfigure}{\columnwidth}
        \centering
        \includegraphics[width=\textwidth]{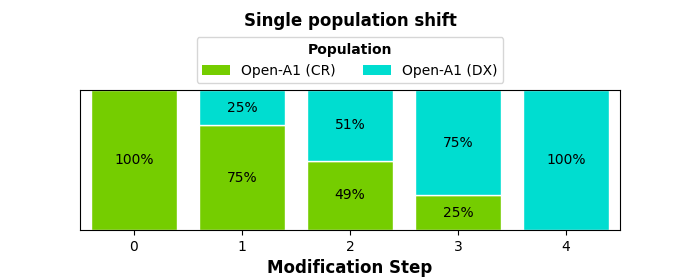}
        \caption{}
        \label{fig:2p-data}
    \end{subfigure}
    \begin{subfigure}{\columnwidth}
        \centering
        \includegraphics[width=\textwidth]{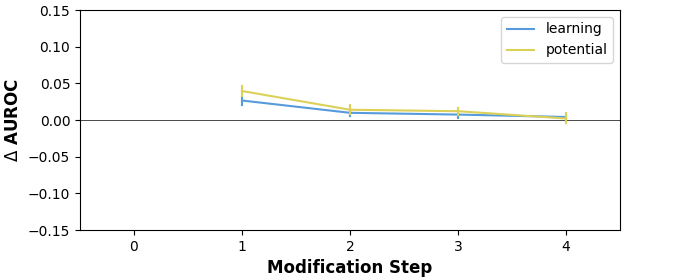}
        \caption{}
        \label{fig:2p-plasticity}
    \end{subfigure}%
    
    \begin{subfigure}{\columnwidth}
        \centering
        \includegraphics[width=\textwidth]{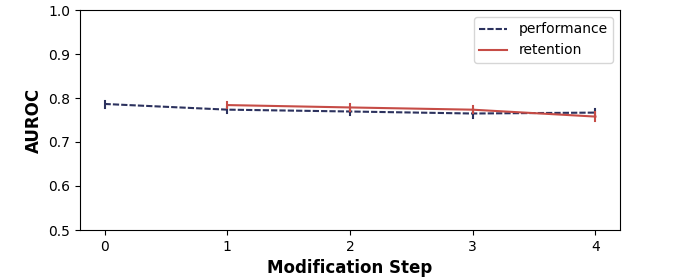}
        \caption{}
        \label{fig:2p-stability}
    \end{subfigure}
    \caption{(a) Population distribution of training, validation, and testing data, (b) \textit{learning} \& \textit{potential} and (c) \textit{retention} \& performance for a model trained and evaluated on a dataset gradually transitioning from one population to another.
    Vertical markers indicate 95\% confidence intervals from models across 25 repetitions.
    }
    \label{fig:results:2p}
    \clearfix
\end{figure}

The first simulated experiment, single population shift, is shown in Figure \ref{fig:results:2p}. The model's performance and retention (Figure \ref{fig:2p-stability}) are relatively stable across all modification steps, as expected due to the gradual transition between the two populations (Figure \ref{fig:2p-data}) which creates a high degree of similarity between each dataset modification step and the previous one. The model's \textit{potential} and \textit{learning} are shown in Figure \ref{fig:2p-plasticity}. The highest \textit{potential} is observed at modification step 1 when the second population, Open-A1 (DX), is first introduced. The model's \textit{learning} closely follows its \textit{potential}, as anticipated due to the combination of uninhibited model plasticity and a gradually transitioning dataset.

\begin{figure}[tb]
    \centering
    \begin{subfigure}{\columnwidth}
        \centering
        \includegraphics[width=\textwidth]{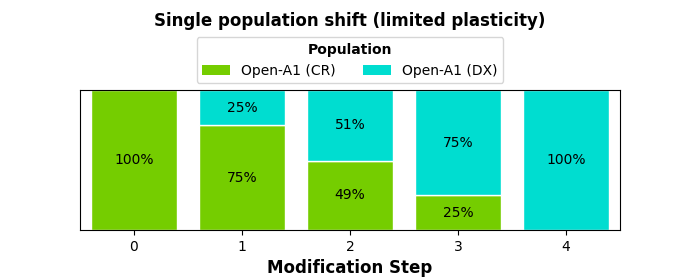}
        \caption{}
        \label{fig:lp-data}
    \end{subfigure}
    \begin{subfigure}{\columnwidth}
        \centering
        \includegraphics[width=\textwidth]{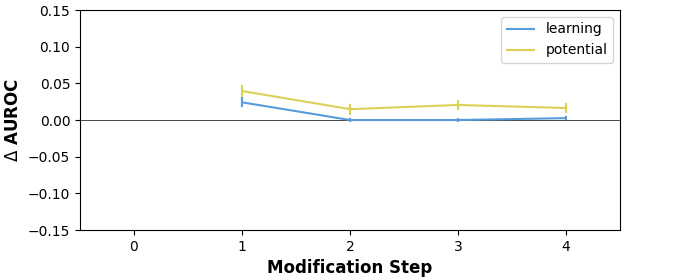}
        \caption{}
        \label{fig:lp-plasticity}
    \end{subfigure}%
    
    \begin{subfigure}{\columnwidth}
        \centering
        \includegraphics[width=\textwidth]{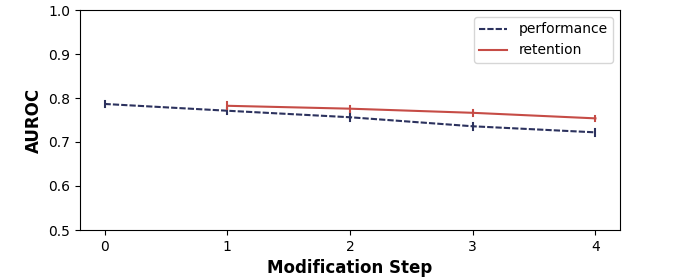}
        \caption{}
        \label{fig:lp-stability}
    \end{subfigure}
    \caption{(a) Population distribution of training, validation, and testing data, (b) \textit{learning} \& \textit{potential} and (c) \textit{retention} \& performance for a model with limited plasticity trained and evaluated on a dataset gradually transitioning from one population to another.
    Vertical markers indicate 95\% confidence intervals from models across 25 repetitions.}
    \label{fig:results:lp}
    \clearfix
\end{figure}

The second simulated experiment, single population shift (limited plasticity), utilizes the same gradually transitioning dataset as the first (Figure \ref{fig:lp-data}). Unlike the previous experiment, a gradual decrease in performance is observed over the four modification steps (Figure \ref{fig:lp-stability}). Analysis of the model's \textit{learning}, \textit{potential} and \textit{retention} reveals the source of this decrease: the model's \textit{learning} never reaches its \textit{potential} at any of the modification steps (Figure \ref{fig:lp-plasticity}). This behavior is expected for this experiment, as the model's plasticity is reduced by freezing all but one layer, reducing the model's ability to adapt to the new population first introduced in modification step 1. Conversely, the model's \textit{retention} (Figure \ref{fig:lp-stability}) remains relatively stable, further supporting the conclusion that the observed decrease in performance is due to limited model plasticity (ability to adapt to new data) rather than a lack of stability (maintained performance on previously seen data). 

\begin{figure}[tb]
    \centering
    \begin{subfigure}{\columnwidth}
        \centering
        \includegraphics[width=\textwidth]{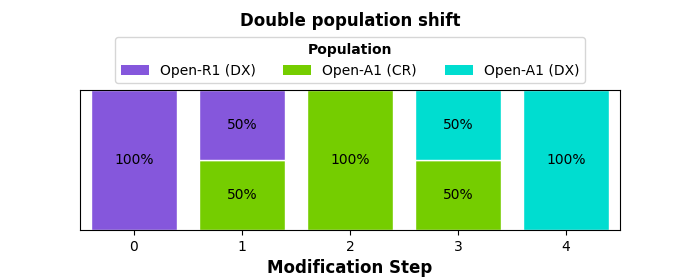}
        \caption{}
        \label{fig:3p-data}
    \end{subfigure}
    
    \begin{subfigure}{\columnwidth}
        \centering
        \includegraphics[width=\textwidth]{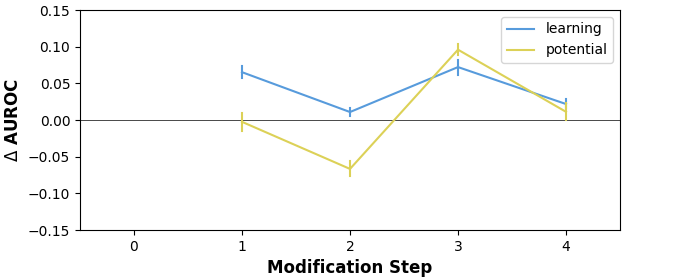}
        \caption{}
        \label{fig:3p-plasticity}
    \end{subfigure}
    
    \begin{subfigure}{\columnwidth}
        \centering
        \includegraphics[width=\textwidth]{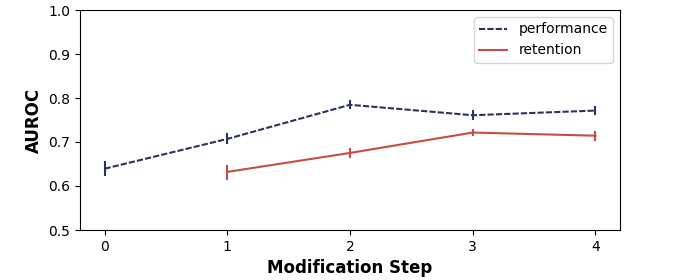}
        \caption{}
        \label{fig:3p-stability}
    \end{subfigure}
    \caption{(a) Population distribution of training, validation, and testing data, (b) \textit{learning} \& \textit{potential} and (c) \textit{retention} \& performance for a model trained and evaluated on a dataset gradually transitioning between three populations.
    Vertical markers indicate 95\% confidence intervals from models across 25 repetitions.}
    \label{fig:results:3p}
\end{figure}

The final simulated experiment, double population shift, involves a more rapidly changing population, transitioning between three datasets across five modification steps (Figure \ref{fig:3p-data}). The model's performance (Figure \ref{fig:3p-stability}) increases over the modification steps, albeit inconsistently. This inconsistency can be explained by the model's \textit{learning} and \textit{potential} (Figure \ref{fig:3p-plasticity}). The model's \textit{potential} (and thus \textit{learning}) spike with the initial introduction of a new population (modification steps 1 \& 3) and dip when receiving additional training on previously seen populations (modification steps 2 \& 4). The lowest \textit{potential} is observed at the same point as the highest performance (modification step 2), suggesting that the high performance was influenced more by the decreased challenge of the population than by improvement to the model. Conversely, the third population -- first introduced at modification step 3 -- presented a greater challenge to the model, as evidenced by the dramatic increase in \textit{potential} and \textit{learning}. This is further supported by observing a decrease in performance accompanied by an increase in \textit{retention}; the increased challenge of the new test dataset results in slightly lower performance while the increased challenge of the new training dataset bolsters performance on previously seen data.

\section{Discussion \& Conclusion}
Traditional performance assessments are designed to measure overall performance by accounting for the expected variability in device inputs and outputs under the assumption of a locked AI model. However, they may be insufficient when applied to adaptive AI systems whose inner workings continually evolve. The adaptive AI paradigm presents opportunities for model improvement, which may be prudent in response to circumstances such as improving compatibility with input acquisition devices, an observed decrease in performance, a known change in the intended population, or adaptation to new technological advances. In this paradigm, simultaneous sequential modifications to the AI model and evaluation datasets introduce multiple axes of change that cannot be captured by a single measurement. As demonstrated in our case study in Section \ref{sec:casestudy} and corresponding results in Section \ref{sec:results}, our proposed measurements may be able to distinguish performance changes due to model adaptations versus dynamic environments with changing data.

The double population shift experiment (Figure \ref{fig:results:3p}) demonstrated that, for systems with dramatically changing populations, the traditional performance assessment does not directly capture the model's retention of previously learned knowledge. Furthermore, the volatility observed in the \textit{learning} and \textit{potential} measurements serve as an indicator of substantial changes to the population, which can suggest a need for increased scrutiny of the model, as the subsequent training may have caused the model to deviate substantially from its initial modification step.

Conversely, the single population shift experiment (Figure \ref{fig:results:2p}) demonstrated that gradual changes to a population presents an opportunity for an AI to learn from the newly introduced data without sacrificing its previously learned knowledge. Additionally, the single population shift (limited plasticity) experiment (Figure \ref{fig:results:lp})  showed that efforts to improve AI knowledge retention come at the cost of the AI's ability to learn new information, resulting in a more stable, but less adaptable, model. 

While our case studies highlight several potential adaptive AI scenarios, these measurements can potentially be applied broadly to disentangle changes in performance due to model plasticity and stability. Table \ref{tab:uses} provides further examples of how each of the measurements may be used to better understand adaptive AI performance.

\begin{table}[htb]
    \small
    \centering
    \begin{tabular}{rp{0.7\linewidth}} \hline
        \textbf{Learning} &  \textit{Measures the effect of model modification with respect to the current evaluation dataset.} \vspace{5pt} \\ 
         & High learning means that the modifications improved model performance for the current population, but it may also indicate a lack of consistency in the model over time. If previous datasets still represent a portion of the current intended population, \textit{retention} may need to be measured. 
         \vspace{3pt}\\ 
         & Negative learning indicates that the current model performs worse than the previous model on the current dataset. If this is observed alongside an increase in performance, the new dataset may merit examination.\\ \hline
    \textbf{Potential} & \textit{Indicates the challenge posed by the new dataset from the perspective of the prior model.}
    \\
     & High potential may indicate the addition of a new population. If a new population was not deliberately added, a population shift may have occurred. \\ \hline
    \textbf{Retention} & \textit{Indicates the model's performance on previous datasets.}
    \\ 
     & Low retention may be a concern when previous datasets represent a portion of the current intended population.\\ \hline
    \end{tabular}
    \caption{Overview of the interpretation and potential utilities of the proposed \textit{learning}, \textit{potential}, and \textit{retention} measurements.}
    \label{tab:uses}
\end{table}

Adaptive AI presents an opportunity for devices to continue to learn after initial deployment, potentially improving device performance and allowing for adaptation to changing environments. However, the evaluation of changing models presents challenges beyond those typically encountered in the evaluation of locked models. In this work, we try to address one such limitation by introducing three measurements -- \textit{learning}, \textit{potential} and \textit{retention} -- which potentially assist with disentangling performance changes attributable to model adaptations versus dataset shifts. Future work may explore the application of these metrics to other medical device AI applications and their potential for integration into automated AI monitoring systems.

\section*{Code Availability}
An implementation of the measurements proposed in this work is available through the VIGILANT project (\href{https://github.com/DIDSR/VIGILANT}{github.com/DIDSR/VIGILANT}), both as a python package and through your browser (\href{https://didsr.github.io/VIGILANT}{didsr.github.io/DIDSR/VIGILANT}).
\section*{Acknowledgments}
This project was supported in part by an appointment to the ORISE Research Participation Program at the Center for Devices and Radiological Health (CDRH), U.S. Food and Drug Administration, administered by the Oak Ridge Institute for Science and Education through an interagency agreement between the U.S. Department of Energy and FDA/Center. 
The imaging and associated clinical data were downloaded from the Medical Imaging and Data Resource Center (MIDRC) and used for research in this publication.
This article reflects the views of the authors and does not represent the views or policy of the U.S. Food and Drug Administration, the Department of Health and Human Services, or the U.S. Government.  The mention of commercial products, their sources, or their use in connection with material reported herein is not to be construed as either an actual or implied endorsement of such products by the Department of Health and Human Services.
\bibliographystyle{unsrt}
\bibliography{references}

\end{document}